# Road Traffic Sign Recognition Method Using Siamese Network Combining Efficient-CNN-Based Encoder

Zhenghao Xi, *Member, IEEE*, Yuchao Shao, Yang Zheng, *Member, IEEE*, Xiang Liu, *Member, IEEE*, Yaqi Liu, and Yitong Cai

*Abstract*— Traffic signs recognition (TSR) plays an essential role in assistant driving and intelligent transportation system. However, the noise of complex environment may lead to motion-blur or occlusion problems, which raise the tough challenge to real-time recognition with high accuracy and robust. In this article, we propose IECES-network which with improved encoders and Siamese net. The three-stage approach of our method includes Efficient-CNN based encoders, Siamese backbone and the fully-connected layers. We firstly use convolutional encoders to extract and encode the traffic sign features of augmented training samples and standard images. Then, we design the Siamese neural network with Efficient-CNN based encoder and contrastive loss function, which can be trained to improve the robustness of TSR problem when facing the samples of motion-blur and occlusion by computing the distance between inputs and templates. Additionally, the template branch of the proposed network can be stopped when executing the recognition tasks after training to raise the process speed of our real-time model, and alleviate the computational resource and parameter scale. Finally, we recombined the feature code and a fully-connected layer with SoftMax function to classify the codes of samples and recognize the category of traffic signs. The results of experiments on the Tsinghua-Tencent 100K dataset and the German Traffic Sign Recognition Benchmark dataset demonstrate the performance of the proposed IECES-network. Compared with other state-of-the-art methods, in the case of motion-blur and occluded environment, the proposed method achieves competitive performance precision-recall and accuracy metric average is 88.1%, 86.43% and 86.1% with a 2.9M lightweight scale, respectively. Moreover, processing time of our model is 0.1s per frame, of which the speed is increased by 1.5 times compared with existing methods.

## I. Introduction

WITH the development of computer technologies, image recognition has been widely applied in smart fields and scenarios such as security defense, imaging medicine and auto driving [1].

Traffic sign recognition (TSR) is an important problem for image recognition. It plays a crucial part in driver assistance systems and autonomous driving systems nowadays [1], [2]. Following regulations with standard shapes and color, traffic signs should be easily detected and recognized by pattern recognition systems in theory. Nevertheless, when the vehicle is driving, images of traffic signs captured by camera on vehicles can be affected by the noise of complex factors from outside environment such as weather conditions, illumination, occluded by leaves or bars and motion artifacts [3], [4]. These non-ideal images hugely increase the difficulty level of the TSR problem, so researchers keep on raising new algorithms to improve the execution efficiency, recognition rate and robustness [5].

In order to solve the problems as above, a novel traffic sign recognition method of Siamese neural network with Convolutional encoder was proposed in this article, and this method is applied to significantly enhance improve recognition rate and robustness of traffic sign recognition. The remarkable ability of encoder and Siamese network have already been revealed by Liu [6], Zhang [7] and chen [8] recently. The main contributions of this paper include the following:

1) Convolutional neural network to extract and encode the traffic sign features of training samples and reference samples.
2) Siamese neural network is designed with Efficient-CNN based encoder and contrastive loss function, which can be trained to improve the robustness of TSR problem when facing the samples of motion-blur and occlusion.
3) Recombined feature code and a fully-connected layer with SoftMax function to classify the codes of samples and recognize the category of traffic signs.

## II. Related Work

Traffic sign recognition technology is mainly composed of traffic sign detection and classification. As for traffic sign detection, traditional methods based on features such as particular color and shape of traffic signs have been widely researched. By converting RGB color space to other color spaces, [9] and [10] use the color threshold to extract focus areas and reduce the interference of illumination.

Received 11 September 2024; revised 25 November 2024; accepted 9 January 2025. This work was supported by the National Natural Science Foundation of China under Grant 12104289. The Associate Editor for this article was S. Mumtaz. *(Corresponding author: Zhenghao Xi.)*

Zhenghao Xi, Yuchao Shao, Xiang Liu, Yaqi Liu, and Yitong Cai are with the School of Electronic and Electrical Engineering, Shanghai University of Engineering Science, Shanghai 201620, China (e-mail: zhenghaoxi@hotmail.com; tongshao66@outlook.com; xliu@sues.edu.cn; ulrich@163.com; yitongcai@163.com).

Yang Zheng is with the Institute of Automation, Chinese Academy of Sciences, Beijing 100190, China (e-mail: yang.zheng@ia.ac.cn).

Digital Object Identifier 10.1109/TITS.2025.3530940

References [6] and [11] detect the shapes of traffic signs using corner detection and Hough Transform of Shape-based methods, respectively. Therefore, combing with color and shape detection, better methods are mentioned such as [12] with color segmentation to locate the signs and shape prior information to check the type of signs. However, color and shape-based methods are unreliable when the traffic signs facing the noise of occlusion, illumination and blur conditions.

As for traffic sign classification, traditional machine learning-based methods are applied to class recognition. For instance, [9] and [13] used support vector machine (SVM) and sparse representations to recognize standard traffic signs, respectively. They also come with the vulnerable performance when facing noise interfered images.

Deep convolutional neural networks, as the modern method, have been rapidly developed for object detection and classification. By training the deep neural network, it can extract image features adaptively because its representation of image features is not fixed. Since Yann LeCun adopt CNN to the traffic sign recognition [14], many approaches based on CNN or its improved networks have been raised in solving the TSR problem, such as R-CNN [2], [15], D-CNN [16]. The images of traffic sign with bounding boxes and labels are deployed as the input of these methods. Other methods such as MPCNN network [17] which transformed from the CNN structure, uses 2D and 3D pixel information for classification neural network to obtain spectral spatial semantic feature recognition of hyperspectral images. These CNN-Based methods lift detection performance, but increase the complexity and size of the model, which slower the processing speed.

There are also quicker methods such as YOLOv3 [18], but it hands poor score on small traffic signs detections. Then, researchers also raise various improved methods based on this fruitful structure. For instance, Akilan et al. [19] present an integrated model that combines multi-view receptive field and encoder-decoder convolutional neural network to detect foreground vehicle objects. A network based on YOLOv3 was proposed by Lee [20] to estimates the location and precise boundary of traffic signs simultaneously. For the small traffic signs detection and recognition, a perceptual generative adversarial network model which narrows the representation difference between small and large traffic signs is showed in [21].

Although much research has been done for TSR, two remaining problems still slower the development of this technology. On the one hand, unlike their superb performance on large or medium traffic signs, the above methods become intensive when facing small traffic signs. On the other hand, complex scene with light and occlusion interference can also give the methods which are trained with moderate samples sharp decline on the accuracy.

To alleviate the problems, Min raises a method [22] based on semantic scene understanding and structural traffic sign location, establishing trusted search regions and analyzed a scene semantically and accurately and then segmented traffic signs in complicated environments precisely. He also uses multiscale densely connected object detector to recognize very small traffic signs. It turns out to be effective with huge amounts of parameters. Another smaller model raised by Wang [23] shows the possibility of good performance with Siamese network and improved R-FCN subclass detector. This method improves detection performance by using locations and sizes of signs as prior knowledge and a lightweight superclass detector. Then, a refinement classifier based on similarity measure learning for subclass classification is constructed to recognize the traffic signs.

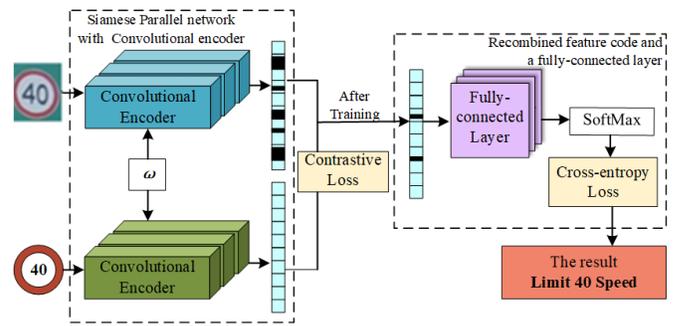

Fig. 1. The framework of IECES-network.

Some contrastive learning method of visual representations can also be used to solve the TSR problem. Such as [24] proposed the SimCLR, this framework can improve the quality of the learned representations. Reference [25] with a combination of normalization and augmentation techniques, using the BYOL achieves the better results in TSR task. But there are lots of Parameters to process and may occupy much computation in these methods.

By contrast, we effectively combine the methods of Siamese [8] and CNN-based network, and propose an improved method with Efficient-CNN based Encoder in Siamese network (IECES-network). Not only can the IECES-network be more robust of the TSR problem when facing the samples of motion-blur and occlusion, but it can show better performance with fewer parameter than the above-mentioned state-of-the-art algorithms.

## III. OUR APPROACH

The framework of our proposed IECES-network is shown in Fig. 1. When inputting an image captured from the on-vehicle camera, we first extract and encode its features with a convolutional encoder based on Efficient-CNN in the first branch. Then we compute the distance of pairs of codes with the input image and template signs with Siamese network to optimize the loss function during the training of encoder. In this way, recombined feature codes of input image can be produced by the improved encoder with closer distance with the templates, which augment the encoder and generate more similar input codes with the template. Finally, the classification step is executed by the improved encoder and a fully-connected layer to output the correct class of the input image.

### A. Coding Sign Features: Convolutional Encoder Based on Efficient-CNN

It's also generally known that the traffic signs are always designed with clear rules and regulations, so that these signs usually presented with prescriptive samples. However, when the traffic signs are captured with the noise of motion-blur and occlusions under different conditions of surroundings, these traffic sign images do not correspond to the prescriptive samples, and very difficult to recognize them. Therefore, we extract the features of traffic signs with simple prior knowledge from these clear regulations. Following this route, we propose a new method to solve the TSR problem with convolutional encoder to encode traffic signs in this section, as shown in Fig. 2. The convolutional encoder has remarkable ability, which was revealed by Liu [26] and Zhang [7].

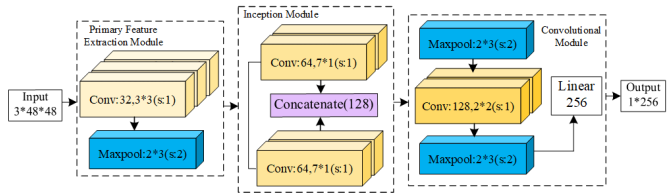

Fig. 2. Convolutional encoder structure.

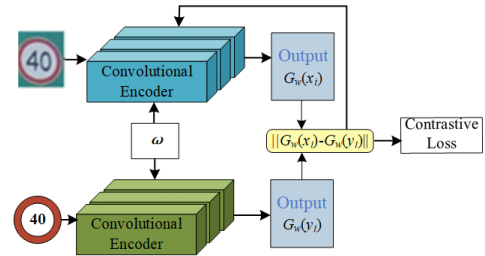

Fig. 3. Siamese neural network convolutional encoder structure.

According to the universal approximation theorem [27], neural networks can approximate any function that maps from any finite dimensional discrete space to another. We choose Efficient-Convolutional Neural Networks (Efficient-CNN) [28] as a mapping function $G_w(\cdot)$, and use $G_w(\cdot)$ to establish a convolutional encoder to extract the feature codes $G_w(I)$ of traffic sign images $I$, $w$ is the parameter of convolutional encoder. This encoder is designed to provide feature codes for classification network in the following three parts: primary extraction layer, inception module [29], [30], [31] and convolutional module.

We first normalize the input traffic sign images as $3 \times 48 \times 48$ samples, 3 is the number of channels, $48 \times 48$ is the resolution ratio of these images.

**Part 1**: In the primary extraction layer, the sample gets through convolutional layers and a max pooling layer. We can get primary features maps contains of the basic information such as edge and color, etc. To ensure each primary features grids' scale is the same for the next layer or module's process, we fulfill the sample's surround with 0 to adjust these grids to the same scale as the input traffic sign image ($3 \times 48 \times 48$). In this way, we receive the $3 \times 48 \times 48$ scale primary features maps.

**Part 2**: We input the primary features maps into the Inception module to extract multi-scale features. This module consists of different scales of convolution kernels based on the Inception V3 network. With the design of asymmetric convolution, through the $3 \times 1$, $1 \times 3$, $7 \times 1$, $1 \times 7$ kernels, we can cascade input images and have the muti-scale features map.

**Part 3**: In the convolutional module. We send the muti-scale features cascade map through the max pooling layers, convolutional layers and a linear fully connected layer to output the 256-bit feature code $G_w(I)$. $G_w(I)$ contains of all crucial information filtered by previous modules.

As above, we introduce a Convolutional encoder module based on Efficient-CNN to extract and encode features. From the output of this part, the input traffic sign images have become codes with multi-scale information.

### B. Computing the Distance: Siamese Parallel Network

To reduce the interference of noise on the feature encoding recognition of traffic signs, we propose a method that uses Siamese network to reduce the difference between the influenced image and templates in the IECES-network. The structure of our Siamese network as shown in Fig. 3. In this method, we compute the distance between the image pairs, and choose the pair of shortest distance as the result.

We add the encoder from Section III. A as a part of the Siamese network and double it, so that we can have two sub-networks with the same structure. We let there are n training samples, the set of training samples is $X = x_1, x_2, \cdots, x_i, \cdots, x_n$, $x_i(i = 1, 2, \cdots, n)$ is any one from $X$ and $G_w(x_i)$ is the feature code of training sample xi. Similarly, we let C categories, the set of standard images (one standard image per category) as $Y = y_1, y_2, \cdots, y_j, \cdots, y_C$, $y_j(j = 1, 2, \cdots, C)$ is any one from $Y$ and $G_w(y_j)$ is the feature code of standard sample $y_i$. Then, we can get a code pair of a training sample and a standard sample $G_w(x_i), G_w(y_j)$.

Also, we have the sharing set of weight $w$. $G_w(x_i), G_w(y_j)$ can be seen as the convolutional code in the comparison space from the encoders $G_w$. Considering the possible blur or occlusion in $X$, we show sample's code pairs $G_w(x_i), G_w(y_j)$ to the mapping function (1). By going through the Section III-A, the image pair should be transformed into code pairs.

$$D_w = \|G_w(x_i) - G_w(y_j)\| \quad (1)$$

The code pairs will be sent to compute the Euclidean distance $D_w$ [32] to check the difference between the sample $x_i$ from $X$ and the sample $y_j$ from $Y$. The contrastive loss function can be defined as follows:

$$L_{sim} = (1-\gamma)D_w^2 + \gamma\left(\max\left(0, m - D_w^2\right)\right) \quad (2)$$

where, $\gamma$ is a variable to judge whether the training sample $X$ belongs to the same category of the standard traffic sign image, and $m$ is the threshold to separate the different categories. When $\gamma = 0$, $x_i$ belongs to the category of $y_j$, otherwise $\gamma = 1$ it's not. The function can be defined as below in details:

$$L_{sim} = \begin{cases} D_w^2 & \gamma = 0 \\ 0 & \gamma = 1, D_w^2 > m \\ m - D_w^2 & \gamma = 1, D_w^2 < m \end{cases} \quad (3)$$

With function (3), we can regard the training process of using the same category as Expectation Maximization (EM) algorithm [8]. During the training, we optimize the encoder of $X$ and $Y$ alternately. Let the clear sample be shown to the network as supervisory information, so that the code of the influenced samples can have shorter distance to the category.

Let $\eta$ becomes the standard code of a category, $\eta_{xi}$ is the standard image code of the category which is $x_i$ belong to. With function (1) and (3), we can get the training loss as follow:

$$L(w, \eta) = \mathbb{E}_{x_i, \tau}\left[\|G_w(\tau(x_i)) - \eta_{xi}\|_2^2\right] \quad (4)$$

where, $\tau(\cdot)$ is a process of augmentation on the training set as Section IV-A. We define the loss of expectation as the expected similarity of interfering sample's code $G_w(\tau(x-i))$ and $\eta_x i$. Then we search for the parameter to minimum the loss of expectation and enable the encoder to exact the correct information of $x_i$ similar to $y_i$ via function (5).

$$\min_{w,\eta} L(w, \eta) = \min_{w,\eta_{xi}} \mathbb{E}_{x,\tau}\left[\|G_w(\tau(x_i)) - \eta_{xi}\|_2^2\right] \quad (5)$$

The optimize of $w$ and $\eta$ will be iterated following function (6) and (7). Where $t$ is the number of iterations:

$$w^t \leftarrow \arg\min_w L\left(w, \eta^{t-1}\right) \quad (6)$$

$$\eta^t \leftarrow \arg\min_\eta L\left(w^t, \eta\right) \quad (7)$$

When iterating $w$, function (6) should be solved by gradient descent and $\eta$ should be fixed. So $y_i$'s encoder don't receive the backward gradient. After $w$'s iteration, function (5) is equivalent to function (8) and can be solved:

$$\eta^t \leftarrow \mathbb{E}_\tau\left[G_{w^t}(\tau(x_i))\right] \quad (8)$$

We get the function (8) when set the expectation on code in function (9) with the code of $y_i$.

$$\eta^t \leftarrow G_{w^t}(y_i) \quad (9)$$

Additionally, we take the standard convolutional code as supervision during the parameter adjustment and get function (10), where $t + 1$ is the next iteration of $t$'s:

$$w^{t+1} \leftarrow \arg\min_w \mathbb{E}_{x,\tau}\left[\|G_w(\tau(x_i)) - G_w(y_i)\|_2^2\right] \quad (10)$$

Enumeration of section headings is desirable, but not required. When numbered, please be consistent throughout the article, that is, all headings and all levels of section headings in the article should be enumerated. Primary headings are designated with Roman numerals, secondary with capital letters, tertiary with Arabic numbers; and quaternary with lowercase letters. Reference and Acknowledgment headings are unlike all other section headings in text. They are never enumerated. They are simply primary headings without labels, regardless of whether the other headings in the article are enumerated.

We can get the parameters by solving function (10) with gradient descent. In that case, we can have robust codes by encoding interfered traffic sign images to have more similar codes with the standard traffic sign images.

Moreover, it's not important whether to pick $x_i$ and $y_j$ from one category, but necessary to have enough positive and negative samples to ensure the discrimination of every category.

We define $W$ as the difference of output, and we judge whether the input traffic sign belongs to the certain class's template or not using $W$. After training the model, we can gather the traffic sign images of certain class closer with shorter distance $D_w$ between their templates and influenced images. In other words, the traing samples belong to some category will have shorter $D_w$ with the standard traffic sign image than those don't belong to it.

### C. Classification: Recombined Feature Code and a Fully-Connected Layer

To classify the codes from traffic signs, we need a classification module assembled with full connected layers based on SoftMax.

Only the code from $x_i$ will be sent to the classification as it's the target of our mission. The classification code $Z$ can be generated via function (11):

$$Z = \omega^* G_w(x_i) + b \quad (11)$$

$\omega$ is the parameter of transformation matrix and $b$ is the bias. We then establish SoftMax function with $Z$ and compute $p_c$, the possibility of each category:

$$p_c = \frac{e^{Z_c}}{\sum_{k=1}^{c} e^{Z_k}} \quad (12)$$

$z_c$ is the $c$ bit of the classification code $Z$, while $z_k$ is one exact bit between the first bit and $c$. By choosing the most possible category from function (12), we can put accurate recognition on the traffic signs.

To train the classification, we compute the cross-entropy loss $L_{class}$ by function (13):

$$L_{class} = -\sum_1^c x_{ic} \ln(p_c) \quad (13)$$

where $x_{ic}$ is the category of the input training sample. We can generate the general loss function L as function (14):

$$L = \alpha L_{sim} + L_{class} \quad (14)$$

We set the hyperparameter $\alpha = 0.1$ to optimize the contrastive loss $L_{sim}$.

### D. Convergence of the Contrastive Loss Function

In order to further analyze and compare the convergence of loss function $L$, we set the sample pair selected by training contains a pair of positive examples $x_i, y_{same}$ and a pair of negative examples $x_i, y_{diff}$. The contrastive loss of positive examples is $L_{sim}^{same}$, $L_{sim}^{diff}$ for the negative examples via function (3) ($L_{sim}^{diff} \neq 0$). When the codes from encoder of the Section III-A satisfy the Constraint 1, the codes are regarded as the valid codes.

*Constraint 1:* $\exists m_{sep} > 0, D_w^{same} + m_{sep} < D_w^{diff}$.

Where, $m_{sep}$ is the boundary for separation, $D_w^{same}$ from the function (1) when the code pairs are positive examples, and $D_w^{diff}$ also from the function (1) when the code pairs are negative examples.

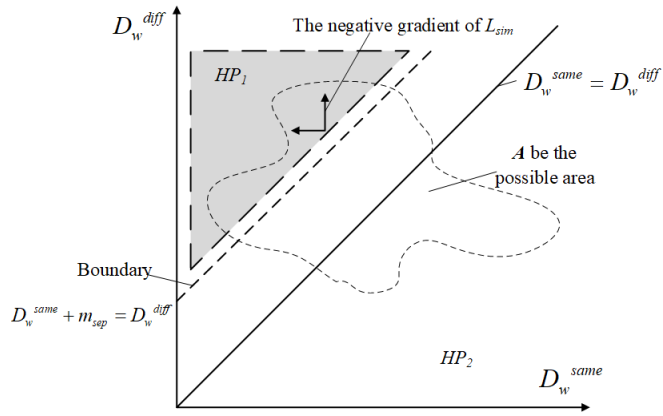

Fig. 4. Plot showing $HP_1, HP_2$ and feasible region $A$.

The contrastive loss function for the positive examples and negative examples can be presented as:

$$L_{sim}\left(D_w^{same}, D_w^{diff}\right) = L_{sim}^{same}\left(D_w^{same}\right) + L_w^{diff}\left(D_w^{diff}\right) \quad (15)$$

When $L_{sim}$ is a convex function on $D_w^{same}$ and $D_w^{diff}$. There are some $w$ meet the Constraint 1. In this case, $L_{sim}$ should satisfy Constraint 2:

*Constraint 2:* The global minimum value $\lambda$ of $L_{sim}\left(D_w^{same}, D_w^{diff}\right)$ is in the half plane of $D_w^{same} + m_{sep} < D_w^{diff}$.

When $\lambda = \infty$, $L_{sim}$ should satisfy Constraint 3:

*Constraint 3:* Positive result exists when the negative gradient of $L_{sim}$ on the boundary of $D_w^{same} + m_{sep} = D_w^{diff}$ have dot product with direction vector $(-1,1)$.

We summarize the above process as a theorem and prove it.

*theorem:* When $L_{sim}\left(D_w^{same}, D_w^{diff}\right)$ is a convex function on $D_w^{same}$ and $D_w^{diff}$, its global minimum value $\lambda$ exsists at infinity. If Constraint 3 is established, there will be a $w$ satisfied Constraint 1.

*Proof:* Let $HP_1$ as the area of $D_w^{same} + m_{sep} < D_w^{diff}$, $HP_2$ as the area of $D_w^{same} + m_{sep} \geq D_w^{diff}$, and $A$ be the possible area of $w$, as shown in Fig. 4.

$$!D_{same}^* = \arg\min L_{sim}\left(D_w^{same}, D_w^{same} + m_{sep}\right) \quad (16)$$

According to the Constraint 3, the negative gradient's direction is towards $HP_1$ for all the spots on the boundary. As $L_{sim}$ is a convex function, when $D_w^{same} + m_{sep} = D_w^{diff}$:

$$L_{sim}(D_{same}^*, D_{same}^* + m_{sep}) \leq L_{sim}(D_w^{same}, D_w^{diff}) \quad (17)$$

There is a spot which has the distance of $e$ from the boundary spot $(D_{same}^*, D_{same}^* + m_{sep})$, its contrastive loss function is $L_{sim}\left(D_{same}^* - e, D_{same}^* + m_{sep} + e\right)$.

We process this spot with first-order Taylor expansion:

$$L_{sim}\left(D_{same}^*, D_{same}^* + m_{sep}\right) + e \begin{bmatrix} \frac{\partial L_{sim}}{\partial D_w^{same}} & \frac{\partial L_{sim}}{\partial D_w^{diff}} \end{bmatrix} \begin{bmatrix} -1 \\ 1 \end{bmatrix} + O(e)^2 \quad (18)$$

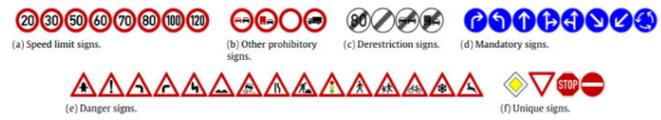

(a) GTSRB database

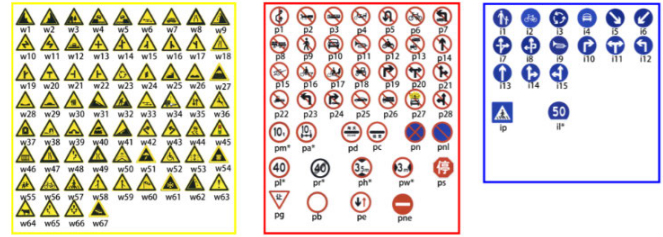

(b) TT100K database

Fig. 5. GTSRB database and TT100K database.

We judge that $e \begin{bmatrix} \frac{\partial L_{sim}}{\partial D_w^{same}} & \frac{\partial L_{sim}}{\partial D_w^{diff}} \end{bmatrix} \begin{bmatrix} -1 \\ 1 \end{bmatrix} < 0$ according to the Constraint 3. In other words, an $e$ can be small enough to satisfy the function (21):

$$L_{sim}\left(D_{same}^* - e, D_{same}^* + m_{sep} + e\right) \leq L_{sim}\left(D_{same}^*, D_{same}^* + m_{sep}\right) \quad (19)$$

As above, a spot can be found in the overlapping area of $A$ and $HP_1$ to let the value of Lsim smaller than any value of Lsim from spots in the overlapping area of $A$ and $HP_2$. According to Constraint 3, there is at least one $w$ satisfy the Constraint 1.

## IV. EXPERIMENT

### A. Experimental Data

In our experiment, we used the German Traffic Sign Recognition Benchmark dataset (GTSRB) [33] and Tsinghua-Tencent 100K dataset (TT100K) [34] such as Fig. 5 to validate the effectiveness of our proposed IECES-network method.

GTSRB contains over 50,000 images of German road signs in 43 classes, where we randomly chose 32,087 images as training samples and 12,629 for the test samples. Then, we set the validation dataset with 20% of training and test samples.

Moreover, we will continuously obtain new segmentation result data by deploying the segmentation method obtained from training the dataset on each terminal of the IoT network. The new segmentation result data continue to supplement our dataset to optimize the model training results. We achieve a closed-loop from "data acquisition", "data processing", to "result sending" to solve the problems of high time cost, slow inference speed, and low segmentation accuracy in the current method using deep learning models to process coal maceral group segmentation.

TT100K is another public dataset contains 10,000 cropped images and 30,000 traffic sign instances from 162 categories. These images cover large variations in illuminance and weather conditions. We used the 2016 annotation of TT100K (containing 42 categories of traffic signs) and randomly chose 19,187 images as training set, 10813 for testing and 20% of all for validation. Taking into account the network compatibility in the experiment, we uniformly cut the input images into the size 48 × 48.

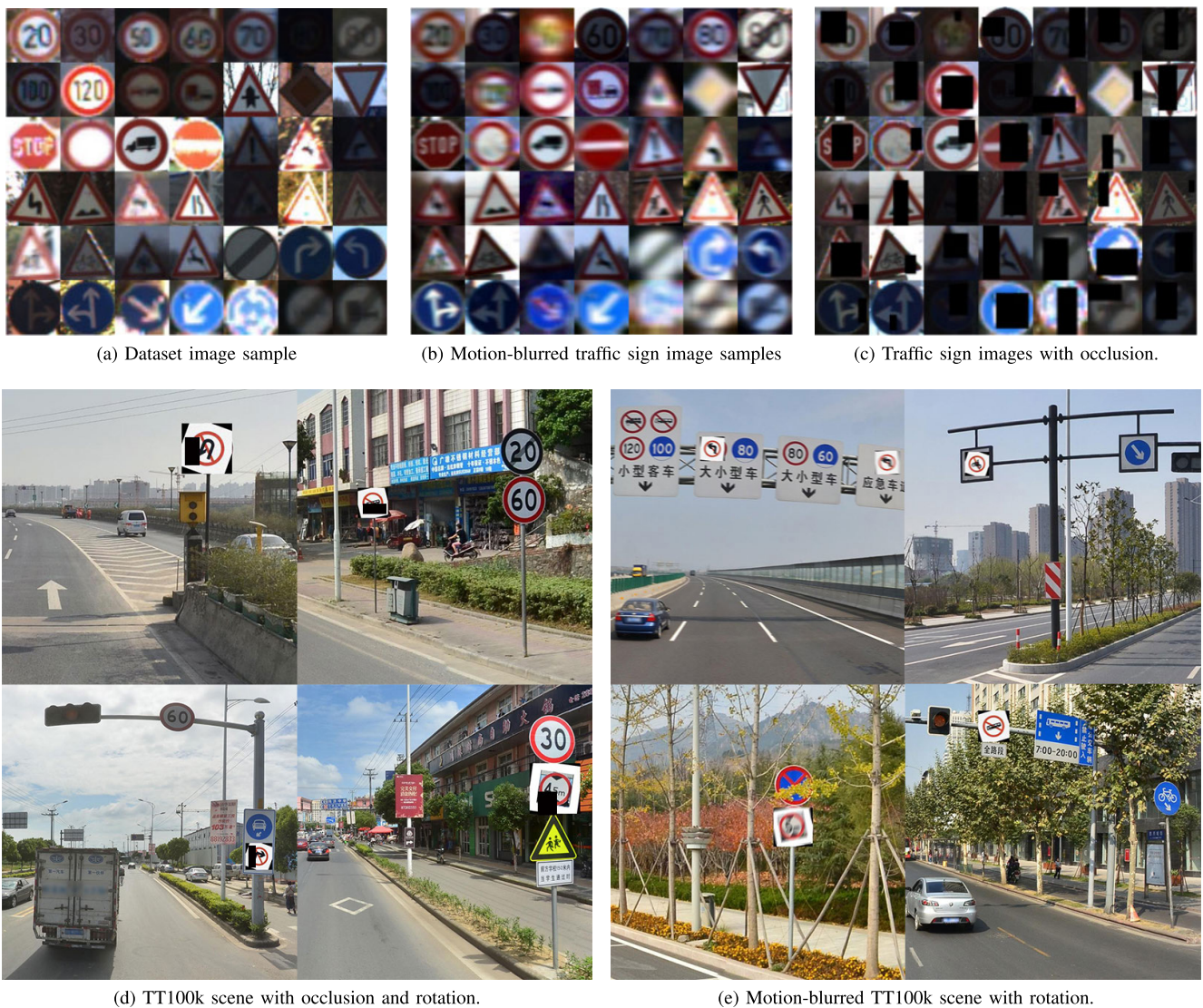

Fig. 6. The results of experimental augmentation.

GTSRB contains over 50,000 images of German road signs in 43 classes, where we randomly chose 32,087 images as training samples and 12,629 for the test samples. Then, we set the validation dataset with 20% of training and test samples.

TT100K is another public dataset contains 10,000 cropped images and 30,000 traffic sign instances from 162 categories. These images cover large variations in illuminance and weather conditions. We used the 2016 annotation of TT100K (containing 42 categories of traffic signs) and randomly chose 19,187 images as training set, 10813 for testing and 20% of all for validation. Taking into account the network compatibility in the experiment, we uniformly cut the input images into the size 48 × 48.

### B. Experimental Augmentation

To simulate the influenced images from vehicle camera, we need to simulate the occlusion and motion-blur scene in the training process so that our proposed IECES-network model can gain the robustness facing these noised images.

To present the occlusion in pictures, according to the fruits from [35], we performed with "Random Erasing". Random Erasing is executed with a certain probability. For an image I in a mini-batch, the probability of it undergoing Random Erasing is p, and the probability of it being kept unchanged is I−p. In this way, training images with levels of occlusion are generated. This effective method randomly selects a rectangle region Ie in an image, and erases its pixels with random values. On the other words, we erase a random rectangle part of training image to stimulate the occlusion scene, shown in the Fig.6c and 6d.

To generate motion-blur scene in training images, we augment both datasets by using the standard image for each class of traffic signs in training set **X** as shown in Fig. 5 and Fig. 6a, adding 5-10 pixels of motion blur to the training sets with an rotation angle in the range [0°, 180°] randomly as motion-blur interference as shown in Fig. 6b and Fig. 6e.

Additionally, for TT100k, we deployed further augmentation [34] by rotating some of its images randomly by an amount in the range [-20°, 20°]. We also scaled it randomly to have size in the range [20, 200] as shown in Fig 6d, and finally add a random but reasonable perspective distortion. In addition, we manually picked TT100K's images without traffic signs and blended in the transformed template, with additional random noise as shown in Fig. 6d and Fig. 6e. This additional process will make TT100k as a more challenging dataset and show the difference between the latest methods and ours.

### C. Evaluation Metrics

To evaluate the recognition performance, we use the precision-recall metric and accuracy. The precision-recall is a curve that measures the trade-off between precision and recall [22]. The equations of these scores are shown as function (20)-(22).

$$\text{Precision (P)} = \frac{TP}{TP + FP} \quad (20)$$

$$\text{Recall (R)} = \frac{TP}{TP + FN} \quad (21)$$

$$\text{Accuracy (A)} = \frac{TP + TN}{TP + FP + TN + FN} \quad (22)$$

where, True Positive (TP) means that estimated sign region has at least 50% overlap with ground truth sign region with sign type being correctly classified. False Positive (FP) means that estimated sign region is incorrectly identified or has no overlapping with ground truth sign region or type. True Negative (TN) means that no identification of traffic sign in non-sign regions. False Negative (FN) means that no identification of traffic sign in sign regions. With these metrics, typical scores are defined to summarize the performance of methods.

### D. Experimental Setting

The experiments are conducted on a workstation with INTEL (R) XEON (R) E5-2680 v4 @2.40GHz and NVIDIA Tesla P40. We implement our IECES-network model by using the PyTorch v2.0 frameworks with CUDA 11.7.512 images from training set will be used as one batch, and all testing model is executed with Adaptive moment estimation optimize (Adam). In our experiment, we set the learning rate of Adam as $3 \times 10^{-4}$, the weight decay as $2 \times 10^{-7}$. Training results will be saved every 20 batches, and the validation of the model will be executed each epoch until the training loss achieved the convergence.

### E. Performance on GTSRB

As our proposed IECES-network model is based on the codes of images, $D_w$ between $G_w(x_i)$ and $G_w(y_i)$ from GTSRB and TT100k becomes essential to the encoder's validity. By generating the heatmap of distance between test image convolutional code and template convolutional code as shown in Fig. 7-9, we can easily find that each distance of same classes is the shortest. These distance below 1.5 can be clearly separated from the distances of different classes, which were above 2.5. For the motion blur images and part-occlusion images, few distances come above 2.5 which as shown in Fig. 9 and 10, because the distances are related to the similarity of pairs of codes. In other words, too serious motion blur and occlusion images may fail to be encoded. In this way, we find that the encoder is strongly robust and effective.

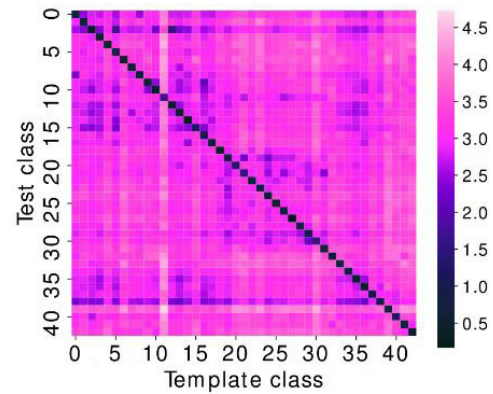

Fig. 7. Heatmap of distance between test image convolutional code and template convolutional code.

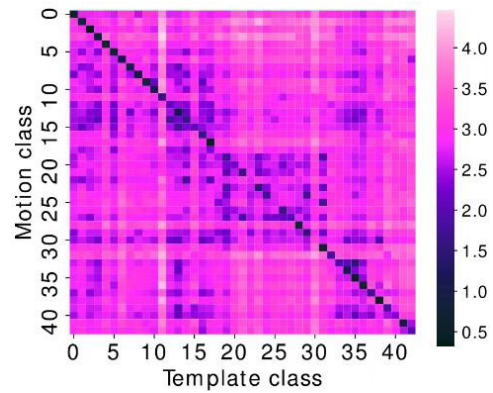

Fig. 8. Heatmap of distance between template convolutional code and convolutional code of traffic sign images with motion-blur interference.

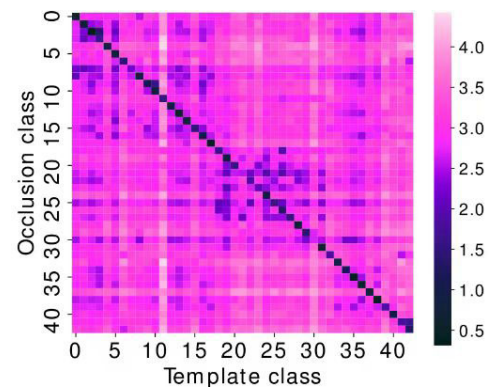

Fig. 9. Heatmap of distance between template convolutional code and convolutional code of traffic sign images with occlusion interference.

In order to highlight the effectiveness our 2.9M proposed network, we conduct dataset testing of five kinds of trained models. To reveal the parameter difference, we pick different scales methods such as Min [22] (34.2M) Zhu [34] (455M), PVT [36] (24.56M), LNL(23.8M) [37] and Wang [23] (6.49M) to compete. With the same settings and input samples, the results came out and are shown in the Table I:

**The comparison on original images: Table II showed the proposed ECES-network mode's accuracy on small samples was 0.1% lower than Min.** For the Prohibitory

TABLE I
THE COMPARISON OF PRECISION-RECALL METRIC FOR THE SOTA METHODS AND OURS ON GTSRB IN DIFFERENT SIZES

| Methods | Zhu(455M) | | Min (34.2M) | | Wang(6.49M) | | PVT(24.56M) [36] | | LNL(23.8M) [37] | | Ours(2.9M) | |
|---|---|---|---|---|---|---|---|---|---|---|---|---|
| | P(Origin) | R(Origin) | P(Origin) | R(Origin) | P(Origin) | R(Origin) | P(Origin) | R(Origin) | P(Origin) | R(Origin) | P(Origin) | R(Origin) |
| Prohibitory | 97.60% | 94.58% | **98.67%** | 96.95% | 93.97% | 87.39% | 97.54% | 96.80% | 96.72% | 94.31% | 98.64% | **97.46%** |
| Danger | 98.42% | 97.94% | 98.56% | 97.89% | 94.77% | 93.09% | 98.04% | 98.13% | 97.23% | 96.41% | **98.98%** | 98.38% |
| Mandatory | 98.01% | 96.26% | 99.18% | **99.65%** | 96.84% | 97.34% | 98.35% | 97.78% | 97.98% | 97.01% | **99.95%** | 97.81% |
| | P(M-B) | R(M-B) | P(M-B) | R(M-B) | P(M-B) | R(M-B) | P(M-B) | R(M-B) | P(M-B) | R(M-B) | P(M-B) | R(M-B) |
| Prohibitory | **95.21%** | 92.78% | 92.51% | 92.42% | 91.08% | 90.32% | 92.23% | 91.48% | 91.56% | 92.35% | 94.86% | **95.59%** |
| Danger | 82.37% | 84.11% | 82.28% | **86.25%** | 81.55% | 80.46% | 81.54% | 85.23% | 81.94% | 82.17% | **91.86%** | 85.47% |
| Mandatory | 87.16% | 88.72% | 91.11% | **91.45%** | 78.21% | 88.61% | 85.19% | 84.93% | 85.18% | 86.72% | **96.74%** | 84.37% |
| | P(Occ) | R(Occ) | P(Occ) | R(Occ) | P(Occ) | R(Occ) | P(Occ) | R(Occ) | P(Occ) | R(Occ) | P(Occ) | R(Occ) |
| Prohibitory | 82.39% | 77.59% | 84.87% | 78.97% | 78.20% | 75.70% | 80.07% | 76.97% | 81.83% | 77.40% | **87.22%** | **81.66%** |
| Danger | 88.53% | 85.22% | 88.55% | 86.60% | 82.97% | 81.89% | 84.69% | 82.67% | 86.63% | 84.95% | **90.88%** | **88.28%** |
| Mandatory | 95.95% | 94.96% | 97.08% | **95.91%** | 90.85% | 92.45% | 92.43% | 91.76% | 95.39% | 93.57% | **98.51%** | 93.55% |

category, IECES-network model gains the best recall and a competitive precision of 98.64%, only 0.03% away from the best result. And the IECES-network model reaches the best precision on Mandatory category, its recall ranks the second with 97.81%. We also report the best accuracy of original samples from GTSDB among four methods.

**The comparison on motion-blur images:** Our IECES-network model gain the 95.59% recall, 2.35% higher than Min's method with the runner-up precision from Table I on the Prohibitory category. For the Danger category, the proposed IECES-network model has taken the best precision and the second recall. IECES-network model gains 5.63% more on precision than the second best, 18.53% more than Wang's method for the Mandatory category.

**The comparison on occluded images:** Table I shows the proposed ECES-network model reaches an outstanding performance on Prohibitory and Danger category. It also achieve the highest precision and a competitive recall on Mandatory category.

In the GTSRB dataset, the reason why the propose IECES-network model could have this dramatic performance of precision-recall metric is because the Siamese network can regard the minor noise from dataset samples as the augmentation to improve the robust of the encoder. When the moderate noise become motion-blur or part-occlusion samples, the precision-recall metric of our IECES-network model is more significant. Other methods have similar performance of precision-recall metric to ours when the interference on samples are minor because they focus on the feature recognition. But they are restricted facing motion-blur and occluded samples.

### F. Performance on TT100K

We evaluate the proposed IECES-network mode on the TT100K dataset to further validate its efficiency and robustness by the accuracy metric. The test image of TT100K dataset contains complex background and different sizes of traffic signs. A typical traffic sign may occupy only 0.2% of the area in images of 2048 × 2048 pixels. It becomes a great challenge to recognize small objects with the complex interference from environment.

According to different traffic sign sizes in pixel, TT100K is divided into three categories: small [0, 32], medium [32, 96], and large [96, 400]. We compare the proposed IECES-network mode with other SOTA methods such as Min [22], Zhu [34], PVT [36], LNL [37] and Wang [23] to evaluate its performance on TT100K dataset. The typical recognition results on TT100K with our method in different case as shown in the Fig. 10.

*1) The Comparison on Original Images:* Table II shows the proposed IECES-network mode's accuracy on small samples was 0.1% lower than Min, the precision and recall takes the second best was 1.6% and 0.9% lower than the best one, respectively. 1.6%, 3% and 1.6% lower with medium samples on accuracy, precision and recall as the runner-up. 2.1%, 0.3%, 2.9% lower with large samples on accuracy, precision and recall also as the second best. With the comparison of each class in Table VII of the Appendix, we find that Min takes the best performance on 40% of classes, while Wang occupies 22.2% and ours is 42.2%. However, Table II shows the proposed IECES-network mode had the smallest parameters which is nearly 45% of Wang and 0.6% of Zhu. Besides, we take the fastest FPS among these methods. Typical recognition results on original images are shown in Table III.

*2) The Comparison on Motion-Blured Images:* Our IECES-network mode achieves the best accuracy on each size. Its accuracy performs 4.8% higher than Min in small samples, 2.8% higher in medium samples, 2.6% and 3.4% higher than Wang in large and overall samples. With the detail comparison, Min performs best on 11.1% of classes, Wang occupies 8.9% and ours takes 82.2% in Table VIII of the Appendix. The precision and recall also takes the best, we demonstrate the robustness from IECES-network mode method has when the input images become more motion noise.

*3) The Comparison on Occluded Images:* The proposed IECES-network mode shows the remarkable result comparing to the others. The overall accuracy reaches 11.4% higher than the second-best method. The lead of the IECES-network mode can beat others in small, medium and large samples at

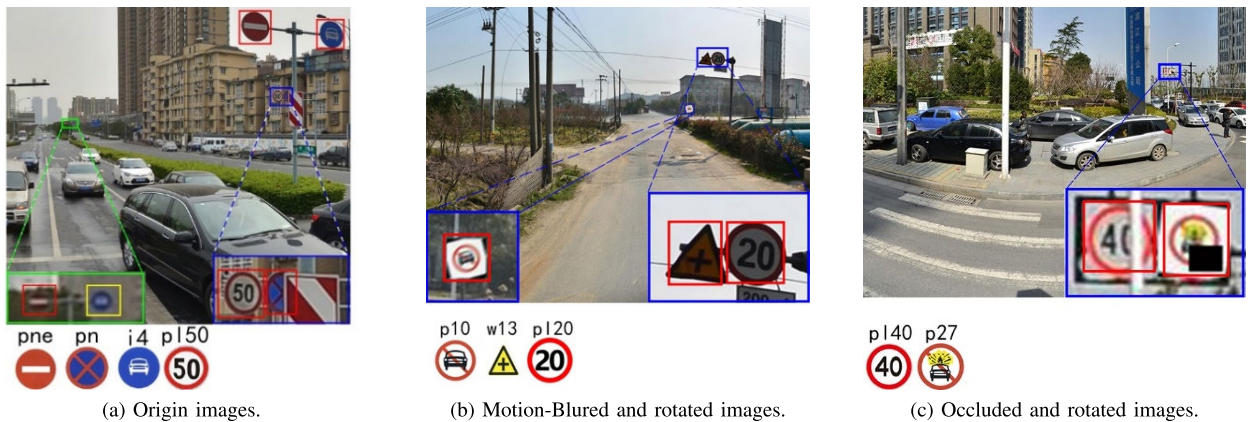

(a) Origin images.   (b) Motion-Blured and rotated images.   (c) Occluded and rotated images.

Fig. 10. Typical recognition results on TT100K with our method in different case.

TABLE II

THE COMPARISON OF PRECISION-RECALL METRIC AND ACCURACY FOR THE SOTA METHODS AND OURS ON TT100K IN DIFFERENT SIZES (M-B FOR MOTION-BLUR AND OCC FOR OCCLUSION)

| Methods | Small(Origin) | | | Medium(Origin) | | | Large(Origin) | | | Overall(Origin) | | |
|---|---|---|---|---|---|---|---|---|---|---|---|---|
| | P | R | A | R | R | A | P | R | A | P | R | A |
| PVT | 83.8% | 82.9% | 84.2% | 92.3% | 92.5% | 92.6% | 92.9% | 93.7% | 93.3% | 89.7% | 90% | 89.2% |
| Min | **89.5%** | **89.2%** | **88.1%** | **96.1%** | **95.5%** | **95.3%** | **95.7%** | **96.1%** | **96%** | **92.1%** | **93.1%** | **92.8%** |
| Zhu | 82.7% | 82.3% | 82% | 91.1% | 91.5%. | 91% | 90.5% | 90.8% | 91% | 87.1% | 87.3% | 87.7% |
| LNL | 85.1% | 84% | 85.4% | 92.5% | 92.7% | 93.1% | 93.4% | 93.1% | 93.6% | 90.4% | 91.2% | 91.5% |
| Wang | 86.4% | 87.1% | 87.3% | 92% | 92.7% | 92.5% | 92.3% | 93.7% | 92.8% | 91.3% | 90.8% | 90.8% |
| Ours | 87.9% | 88.3% | 88% | 93.1% | 93.9% | 93.7% | 94% | 93.2% | 93.9% | 90.1% | 90.4% | 90.9% |
| Methods | Small(M-B) | | | Medium(M-B) | | | Large(M-B) | | | Overall(M-B) | | |
| | P | R | A | R | R | A | P | R | A | P | R | A |
| PVT | 80.1% | 79.5% | 78.2% | 84.3% | 85.5% | 85.4% | 85.7% | 85.9% | 85.3% | 82.6% | 83.4% | 83.3% |
| Min | 80% | 80.7% | 81.1% | 86.6% | 87.3% | 86.5% | 87.1% | 87.5% | 87.9% | 84.1% | 82.3% | 84.6% |
| Zhu | 75.1% | 74.7% | 75.4% | 81.4% | 80.3% | 81% | 83.7% | 83.1% | 82.8% | 78.9% | 78.1% | 79.1% |
| LNL | 77% | 76.1% | 77.1% | 84.4% | 83.2% | 84.6% | 86.1% | 86.4% | 86.6% | 83.5% | 84.2% | 84.5% |
| Wang | 81.4% | 80.9% | 80.6% | 87.3% | 88.2% | 88.1% | 87.1% | 87.6% | 89.5% | 86.1% | 85.9% | 85.2% |
| Ours | **85.8%** | **85.1%** | **85.9%** | **90.4%** | **90%** | **90.9%** | **91.4%** | **91.1%** | **92.1%** | **88.7%** | **88.3%** | **88.6%** |
| Methods | Small(Occ) | | | Medium(Occ) | | | Large(Occ) | | | Overall(Occ) | | |
| | P | R | A | R | R | A | P | R | A | P | R | A |
| PVT | 67.5% | 68.1% | 68.4% | 79.3% | 79.9% | 80.3% | 80.8% | 80.1% | 81.9% | 75.1% | 74.9% | 75.3% |
| Min | 76.4% | 77.5% | 77.7% | 81% | 80.9% | 81.6% | 81.2% | 80.3% | 82.6% | 79% | 79.1% | 79.5% |
| Zhu | 68.5% | 69.4% | 70.1% | 77.1% | 77.8% | 77.9% | 79.2% | 78.5% | 79.3% | 74.5% | 74.7% | 75.1% |
| LNL | 63.1% | 63.8% | 64.3% | 74.2% | 73.7% | 74.7% | 75.6% | 75.4% | 76.8% | 72.1% | 73.8% | 73.9% |
| Wang | 78% | 77.4% | 78.2% | 79.1% | 78.5% | 80.4% | 83.1% | 84.2% | 84.3% | 78.3% | 79.6% | 80.1% |
| Ours | **83%** | **83.2%** | **84.3%** | **86.5%** | **85.4%** | **86.7%** | **85.3%** | **86.1%** | **86.8%** | **85.1%** | **84.7%** | **85.3%** |

precision, accuracy and recall metrics. When it comes to the detail comparison as shown in Table IX of the Appendix, our method takes SOTA performance on 84.4% classes. Typical recognition results on occluded images are shown in Table IV.

Analyze the comparative experimental results on the TT100K dataset. The IECES-network mode also uses the constraints of spatial information relationships, but it has a lower parameter to improve its performance than Min [22].This is because the proposed IECES-network mode not only optimized the encoder to produce more similar codes of these samples to the standard but shortened the distance between all samples in one category as well. It can further the generalizability of standard samples and get a minor distance of samples in one category, and demonstrate the robustness and effectiveness of the IECES-network mode.

The prior knowledge of the proposed IECES-network mode is mainly reflected in the noise environment, the prior knowledge could ensure that the possible class of the sign was set with a narrow range, the process could be more efficient and achieve a better score. With the help of optimized encoder and the shortened code pairs, we could have the least performance degradation when the input samples become more and more complex and occluded and chase closely on the SOTA performance on the moderate samples.

TABLE III
THE COMPARISON OF PARAMETERS AND COST TIMES FOR THE SOTA METHODS AND OURS ON TT100K

| Methods | Parameters | Times(s) |
|---|---|---|
| PVT | 24.56M | 0.105 |
| Min | 34.2M | 0.413 |
| Zhu | 455M | 2 |
| LNL | 23.8M | 0.577 |
| Wang | 6.49M | 0.15 |
| Ours | **2.9M** | **0.1** |

TABLE IV
THE TYPICAL RECOGNITION RESULTS OF MOTION-BLURED IMAGES FOR THE SOTA METHODS AND OURS ON TT100K

| | Min | Wang | Zhu | PVT | LNL | Ours |
|---|---|---|---|---|---|---|
| P12 | P12 | P12 | P26 | P6 | P12 | P6 |
| W57 | W57 | W55 | W57 | W55 | W57 | W57 |
| Pm55 | Pm55 | Pm55 | Pm55 | Pm55 | Pm55 | Pm55 |

TABLE V
THE TYPICAL RECOGNITION RESULTS OF OCCLUDED IMAGES FOR THE SOTA METHODS AND OURS ON TT100K

| | Min | Wang | Zhu | PVT | LNL | Ours |
|---|---|---|---|---|---|---|
| I5 | I5 | I5 | I5 | I5 | I5 | I5 |
| W57 | W57 | W55 | W57 | W55 | W57 | W57 |
| PL120 | PL100 | PL20 | PL40 | PL100 | PL100 | PL120 |

TABLE VI
TRAFFIC SIGN RECOGNITION ACCURACY WITH DIFFERENT ENCODER

| Encoder | Origin accuracy | Occluded accuracy | Motion-blur accuracy | Epoch |
|---|---|---|---|---|
| STN | 98.05 | 86.76 | 90.34 | 2000 |
| Com-CNN | 97.14 | 85.57 | 88.87 | 4000+ |
| Efficient-CNN | 98.40 | 88.57 | 90.52 | 3000 |
| Ours | 98.76 | 90.05 | 92.74 | 4000 |

### G. Ablation Study and Robustness Analysis

*1) Ablation Studies on Training Speed:* To more intuitively demonstrated the better performance of the proposed method, we conducted the ablation study and the results are shown as Table V.

In order to highlight the effectiveness our proposed network, we conduct dataset testing of four kinds of trained models. As the parameter of our work is 2.9M, we pick a 14.6M parameter model (Spatial Transformer Networks [38], STN [39]) and regard it as a large model to compare. Also, we choose the Committee of CNNs (Com-CNN) [40] as a small one because of its scale of 1.5M parameter. Efficient-CNN [28] of 2.8M parameter is trained as the similar size model of ours. With the same settings and input samples, the results come out and are shown below:

As observed from train accuracy line graph of all tested models, each method can achieve the accuracy above 98% with enough epochs except Com-CNN. STN reaches 98.05% with only 2000 iterations while the proposed method needs 4000. Efficent-CNN takes similar performance with 3000 iterations. It means the proposed method needs more positive and negative samples to seek suitable optimized parameters and ensure the encoder's performance.

As for the Loss functions, the proposed method uses contrastive loss function to enhance the encoder and the cross-entropy loss function to optimize the classification layer, presented a lower speed than Efficient-CNN, which includes single optimizing task. But it is the contrastive loss function and a cross-entropy loss function contributes to the performance when the input samples become motion-blurred or occluded. Although there are more iteration times, our accuracy still ranked at a high level of all methods.

*2) Robustness Analyzes on Occluded Samples:* To more intuitively demonstrate the better performance of the proposed method on samples with different level of occlusion, we conduct the ablation study and the results are shown as Table VI, where level P is the proportion of randomly occluded parts in the image.

We replace the moderate test set with different levels of occlusion augmented on images to compare the performance on these samples. When the possibility and degree add up, the proposed method performs better than others and seeks a difference of at most 16% in occlusion test, which as shown Fig. 11.

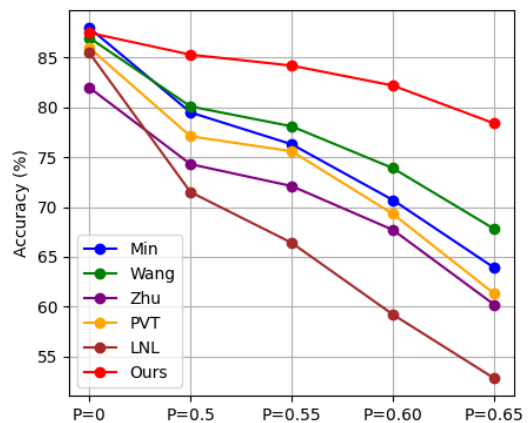

Fig. 11. Train accuracy convergence curve of algorithms with different P.

TABLE VII
THE CASE OF ORIGIN IMAGES

| Class | i2 | i4 | i5 | 100 | II60 | II80 | Ip | P10 | P11 | P12 | P19 | P23 | P26 | P27 |
|---|---|---|---|---|---|---|---|---|---|---|---|---|---|---|
| Min | **88.8** | **96.3** | 95 | 99 | **98.6** | 98.1 | 91.7 | 86.1 | **93.7** | **97.8** | 98 | 97.3 | **95.3** | **99.5** |
| Wang | 85.6 | 93.3 | 97.2 | 94.5 | 95.5 | **98.7** | 90.6 | 83.2 | 90.9 | 94.9 | 94.3 | **98.5** | 93.6 | 97.8 |
| Zhu | 82.3 | 87.1 | 90.5 | 87.7 | 92.1 | 91.5 | 86.1 | 79.4 | 87.5 | 90.3 | 91.7 | 89.2 | 86.1 | 93.9 |
| PVT | 82.1 | 90 | 88.5 | 92.5 | 92.4 | 91.1 | 84.8 | 79.2 | 87.1 | 90.9 | 91.9 | 91.1 | 89.3 | 93.2 |
| LNL | 86.4 | 94.5 | 93.7 | 97.7 | 97.1 | 96 | 89.4 | 84.5 | 90.8 | 95.3 | 95.9 | 95.1 | 93.5 | 98 |
| ours | 87.3 | 94.3 | **96.2** | **100** | 97.84 | 98.5 | **95.2** | **88.3** | 91.2 | 95 | 95.3 | 97.1 | 91.8 | 95.1 |

| Class | P3 | P5 | P6 | pg | Ph4 | Ph4.5 | Ph5 | PL100 | PL120 | PL20 | PL30 | PL40 | PL5 | PL50 |
|---|---|---|---|---|---|---|---|---|---|---|---|---|---|---|
| Min | 93.7 | **94.9** | **92.4** | **95.3** | 91.5 | **89.3** | **85.1** | 99 | 99.6 | 90.4 | 94.2 | **95.6** | **94.2** | **95.1** |
| Wang | 89.1 | 91.5 | 91.6 | 90.1 | **92.2** | 85.9 | 83.2 | 93.8 | **100** | 91.2 | **95.3** | 91.1 | 95 | 91.8 |
| Zhu | 86.4 | 86.9 | 85.2 | 88.5 | 86.0 | 83.8 | 79.9 | 91.4 | 92.8 | 83.3 | 89.1 | 87.7 | 86.7 | 89.5 |
| PVT | 87.3 | 88.6 | 85.6 | 88.9 | 85.2 | 82.8 | 79 | 92.2 | 93.5 | 83.4 | 87.4 | 89.4 | 88.2 | 88.3 |
| LNL | 92 | 92.4 | 91.4 | 94.1 | 90.4 | 88.2 | 82.4 | 96.6 | 97.7 | 89.2 | 92.2 | 93.7 | 92.9 | 93.2 |
| ours | **94.2** | 92.6 | 91.6 | 93.5 | 89.3 | 82.1 | 82.9 | 93.2 | **100** | 90.4 | 90.8 | 95.3 | 93.9 | 92.7 |

| Class | PL60 | Pl70 | Pl80 | Pm20 | Pm30 | Pm55 | pn | pne | Pr40 | W13 | W32 | W55 | W57 | W59 |
|---|---|---|---|---|---|---|---|---|---|---|---|---|---|---|
| Min | 96.3 | **96.1** | **95.8** | 91.7 | **93.2** | **95.7** | **95.6** | **96.8** | 99.8 | **89.2** | **93.9** | **96.4** | **95.1** | 90.5 |
| Wang | 96.9 | 92.3 | 91.7 | **92.2** | 91.4 | 94.3 | 90.9 | 94.6 | 95.6 | 89.8 | 89.9 | 91.3 | 92.2 | 86.1 |
| Zhu | 90.8 | 90.5 | 89.9 | 85.1 | 86.9 | 89.8 | 88.8 | 91.4 | 93.7 | 82.8 | 86.5 | 90.8 | 88.6 | 83.7 |
| PVT | 89.6 | 89.4 | 89 | 85.6 | 86.8 | 89.6 | 88.7 | 90.2 | 93.7 | 82.9 | 87.6 | 89.7 | 88.5 | 83.6 |
| LNL | 94.5 | 93.9 | 93.5 | 90.6 | 91.5 | 93.4 | 93.6 | 94.1 | 98.5 | 88.1 | 91.6 | 95.3 | 92.9 | 87.6 |
| ours | **97.2** | 91.7 | 92.64 | 90.8 | 89.5 | 93.7 | 91.5 | 92.8 | 96.8 | **90.1** | 90.3 | 93.5 | 92.4 | **90.7** |

TABLE VIII
THE CASE OF MOTION-BLURED IMAGES

| Class | i2 | i4 | i5 | 100 | II60 | II80 | Ip | P10 | P11 | P12 | P19 | P23 | P26 | P27 |
|---|---|---|---|---|---|---|---|---|---|---|---|---|---|---|
| Min | 77.9 | 84.5 | **87.9** | 92.1 | 92.3 | 89.1 | 81.4 | 80.8 | 83.2 | 84.1 | 89.2 | 89.2 | 86.5 | 92.4 |
| Wang | 77.4 | 85.2 | 84.3 | 87.6 | 92.1 | 90.5 | 79.8 | 77.6 | 85.1 | 85.2 | 87.5 | 91.4 | 87.6 | 89.2 |
| Zhu | 73.4 | 79.7 | 82.4 | 87.6 | 88.1 | 84.5 | 75.5 | 75.2 | 77.9 | 78.4 | 83.6 | 84.8 | 80.7 | 87.3 |
| PVT | 75.3 | 83.2 | 86.9 | 91.5 | 90.3 | 86.9 | 79.6 | 79.7 | 81.1 | 83.2 | 87 | 87.7 | 83.7 | 91.6 |
| LNL | 76.3 | 81 | 87.3 | 88.7 | 90.8 | 88.2 | 78.7 | 77.2 | 80.1 | 80.8 | 87.7 | 88.1 | 83.4 | 89.1 |
| ours | **84.2** | **87.6** | 87.5 | **97.2** | **94.3** | **97.9** | **93.5** | **86.2** | **87.4** | **92.9** | **92.2** | **95.3** | **89.3** | **95.1** |

| Class | P3 | P5 | P6 | pg | Ph4 | Ph4.5 | Ph5 | PL100 | PL120 | PL20 | PL30 | PL40 | PL5 | PL50 |
|---|---|---|---|---|---|---|---|---|---|---|---|---|---|---|
| Min | 85.7 | 87.8 | 82.3 | 85.2 | 86.1 | **84.5** | **79.1** | **93.3** | 92.1 | 83.7 | 85.2 | 86.2 | 82.9 | 87.8 |
| Wang | 85.2 | 86.3 | 82.6 | 84.7 | 84.3 | 79.2 | 78.5 | 92.4 | 90.8 | 84.2 | 86.5 | 87.8 | 85.3 | 86.2 |
| Zhu | 80.1 | 82 | 77.7 | 81 | 81.6 | 79.6 | 73.5 | 87.6 | 88.1 | 78.7 | 80.4 | 81.8 | 78.7 | 83.1 |
| PVT | 84.3 | 87 | 79.5 | 82.5 | 85 | 82.4 | 76.6 | 91.4 | 90.3 | 82.6 | 84.5 | 83.5 | 80.1 | 85.7 |
| LNL | 81.7 | 85.9 | 80.5 | 82 | 84.4 | 80.7 | 75.6 | 91.3 | 89 | 80.6 | 84.3 | 82.5 | 80.6 | 84.4 |
| ours | **90.3** | **89.6** | **88.1** | **90** | 85.1 | 79.3 | 78.9 | 90.1 | **97.8** | **87.2** | **87.9** | **92.7** | **91.9** | 87.5 |

| Class | PL60 | Pl70 | Pl80 | Pm20 | Pm30 | Pm55 | pn | pne | Pr40 | W13 | W32 | W55 | W57 | W59 |
|---|---|---|---|---|---|---|---|---|---|---|---|---|---|---|
| Min | 89.2 | **90.5** | 86.3 | 85.2 | 86.841 | 87.2 | 87.6 | 86.2 | 92.3 | 84.8 | 86.1 | 85.9 | 87.2 | 81.3 |
| Wang | 90.7 | 86.42 | 84.2 | 86.8 | 85.641 | 91.1 | **89.3** | 85.9 | 93.1 | 86.9 | 84.7 | 83.3 | 88.5 | 82.9 |
| Zhu | 83.3 | 85.9 | 81.3 | 79.8 | 82.1 | 81.3 | 81.7 | 81.7 | 87.7 | 80.2 | 82 | 80.7 | 82.2 | 77.2 |
| PVT | 87.9 | 87.7 | 85.2 | 84.3 | 85.1 | 84.2 | 86.5 | 84 | 91.2 | 82.5 | 84.7 | 83.8 | 85.1 | 79.5 |
| LNL | 87.6 | 86.9 | 84.4 | 84.7 | 83.2 | 86.4 | 86 | 82.4 | 89.8 | 82.1 | 84 | 84.4 | 85.5 | 78.4 |
| ours | **92** | 87.4 | **89.4** | **90.5** | **88.32** | **92.8** | 88.1 | **91.3** | **94.4** | **88.3** | **88.9** | **91.1** | **91.7** | **86.4** |

The proposed IECES-network and Wang, which uses parallel network performs better than Min's method because the Siamese network structure can provide a more robust performance on datasets with intra-class and inter-class data imbalance. Wang [23] provides accurate result on subclasses such as prohibitory, danger, mandatory. But when facing the influenced images, it can't provide sound performance on judge the exact category within these subclasses. The proposed method, which achieves the best performance, using convolutional encoder to transfer samples into codes and compare the distance. With the nearest code pairs of occlusion samples and standard signs, the proposed method can classify the samples more accurately and also provides strong robustness on these situations.

TABLE IX
THE CASE OF OCCLUDED IMAGES

| Class | i2 | i4 | i5 | 100 | II60 | II80 | Ip | P10 | P11 | P12 | P19 | P23 | P26 | P27 |
|---|---|---|---|---|---|---|---|---|---|---|---|---|---|---|
| Min | 74.3 | 80.2 | 82.3 | 86.5 | 88.4 | 84.2 | 75.9 | 76.3 | 80.2 | 82.3 | 85.1 | 86.7 | 81.4 | 88 |
| Wang | 74.6 | 81.2 | 80.5 | 85.4 | 82.1 | 85.4 | 76.9 | 74.3 | 81.9 | 80.4 | 82.3 | 84.8 | 81.5 | 83.3 |
| Zhu | 69.1 | 74.4 | 76.7 | 82.6 | 82.6 | 80.3 | 69.9 | 70.7 | 73.8 | 76.7 | 79.8 | 83.3 | 76.9 | 83.9 |
| PVT | 73 | 75.5 | 77 | 83.2 | 82.7 | 83 | 68.3 | 72.4 | 78.2 | 77.1 | 77.9 | 84 | 75.4 | 86.3 |
| LNL | 67.2 | 75 | 74.8 | 79.4 | 83.2 | 78.9 | 69.8 | 70.2 | 72.8 | 74.5 | 77.1 | 79.4 | 75.3 | 82.7 |
| ours | **83.1** | **84.7** | **84** | **94.82** | **90.2** | **95.4** | **90.7** | **83.2** | **82.1** | **89.9** | **88.5** | **91.4** | **86.5** | **93** |

| Class | P3 | P5 | P6 | pg | Ph4 | Ph4.5 | Ph5 | PL100 | PL120 | PL20 | PL30 | PL40 | PL5 | PL50 |
|---|---|---|---|---|---|---|---|---|---|---|---|---|---|---|
| Min | 79.7 | 81.6 | 77.6 | 79.8 | 82.4 | **79.2** | 75.7 | 86.1 | 87.4 | 78.5 | 79.9 | 81.4 | 77.2 | 83.9 |
| Wang | 79.7 | 82.8 | 75.3 | 804 | **81.7** | 76.9 | 74.2 | 84.9 | 86.3 | 82.9 | 81.2 | 83.5 | 80.1 | 84.2 |
| Zhu | 75.7 | 74.7 | 73 | 73.2 | 76.9 | 73 | 70.7 | 80.8 | 82.4 | 74.7 | 74 | 77.3 | 74.1 | 78.3 |
| PVT | 72.2 | 74.1 | 69.7 | 71.5 | 77.3 | 74.2 | 68.3 | 79.9 | 80.8 | 71.1 | 71.8 | 77.7 | 73.2 | 82.1 |
| LNL | 72.4 | 74.9 | 71.3 | 72.1 | 77.1 | 72.7 | 70.7 | 79.7 | 82.2 | 73.1 | 74.5 | 74.5 | 70 | 77.1 |
| ours | **87.4** | **84.9** | **86.3** | **87.5** | 81.3 | 76.1 | **77.3** | **88.2** | **95.3** | **84.9** | **85.8** | **89.6** | **87.2** | 81.7 |

| Class | PL60 | Pl70 | Pl80 | Pm20 | Pm30 | Pm55 | pn | pne | Pr40 | W13 | W32 | W55 | W57 | W59 |
|---|---|---|---|---|---|---|---|---|---|---|---|---|---|---|
| Min | 86.7 | 81.2 | 78.6 | 74.9 | 80.5 | 82.6 | 80.1 | 81.2 | 83.6 | 79.8 | 81.3 | 80.2 | 80 | 69.8 |
| Wang | 86.9 | 81.8 | 80.1 | 79.9 | 81.3 | 86.1 | **86.7** | 80.7 | 85.2 | 81.8 | 79.4 | 76.9 | 84.3 | 75.2 |
| Zhu | 83 | 74.4 | 71.8 | 68.2 | 76 | 79.5 | 73.4 | 76.5 | 76.7 | 73.4 | 77.1 | 75.7 | 73.6 | 65.5 |
| PVT | 81.1 | 75.7 | 70.1 | 68.3 | 74.9 | 80.8 | 74.2 | 72.3 | 78.5 | 71.8 | 74.4 | 73.6 | 73.4 | 65.9 |
| LNL | 78.8 | 75.1 | 72.7 | 67.3 | 74.8 | 74.7 | 75.1 | 73.3 | 75.9 | 73.2 | 73.3 | 75 | 73.3 | 61.9 |
| ours | **89.3** | **83.2** | **86.1** | **86.8** | **85.4** | **89.2** | 84.5 | **89.1** | **91.3** | **86.7** | **84.2** | **87.8** | **87** | **81.5** |

As for the uniformity-tolerance dilemma of the contrastive loss, our encoder processes the tolerance pretty well by the proper learning of distance with the same class while maintaining the uniformity to separate different classes of signs at SOTA level. In addition, [8] has researched the ablation problem of Siamese network, we will not elaborate further in this article.

In addition, compared with other Siamese-like methodologies 6, for example, there have been several popular strategies for preventing Siamese networks from collapsing. SimCLR relies on negative samples to prevent collapsing rather than stop-gradient. Dissimilarity ways may occupy much computation and operate on a lower speed. With the core attribution provided by Siamese networks, the minor parameters way is the stop-gradient method. BYOL takes only on positive pairs with momentum encoder used to the same effect, which is another solution towards function 6 and 7 in section B III, to optimize $\eta$, which means that the general CNN encoders can also take the burden in Siamese networks, as ours done. In conclusion, Siamese networks are natural and effective tools for modeling invariance, and the minor parameter approach we adopt demonstrate its effectiveness while contributing to the light-weight methods on the focus of representation learning.

## V. CONCLUSION

In order to further increase the precision-recall and accuracy metric of traffic signs recognition in the motion-blurred and occluded images' scenes, this article proposed a novel traffic sign recognition approach by combining Efficient-CNN based encoders with Siamese network and fully-combined layers. With light Efficient-CNN based encoders which introducing primary extraction layer and Inception model at first, the more useful multi-scale feature cascade map was gained. Then, we designed the IECES-network with contrastive loss function, which could optimize the encoder of inputs images by computing the distance between inputs and templates. Next, the refinement classifier with recombined feature code was proposed, it retained the effectiveness of a simple structure.

Experimental results indicated that the precision-recall and accuracy metric of the IECES-network proposed in this article increased obviously, and especially facing the environment with the noise of motion-blur and occlusion. Moreover, our proposed network costed less parameters and times than other SOTA methods. As above, the advantages of the IECES-network could be mainly concluded as three aspects: high recognition ability, less computation, and good robustness. Based on the important roles played by TSR in assistant driving and intelligent transportation research, it was assured that our research could promote the development of these areas, a light and fast model convenience to mobile and embedded devices. In the future, we would like to fuse more prior elements from other kinds of sensors into our system to achieve superior performance.

APPENDIX

The detail comparison of each class for the SOTA methods and ours on TT100K in different case as shown in the Table VII, VIII and IX.

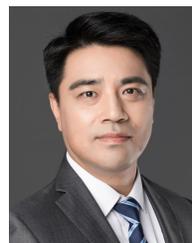

**Zhenghao Xi** (Member, IEEE) received the B.E. degree from the South-Central University for Nationalities, Wuhan, China, in 2003, the M.E. degree from the University of Science and Technology, Liaoning, Anshan, China, in 2008, and the Ph.D. degree from the University of Science and Technology at Beijing, Beijing, China, in 2015.

He is currently a Professor with Shanghai University of Engineering Science, Shanghai, China, and the Director of the Department of Automation. His research interests include computer vision, robotics, and intelligent systems.

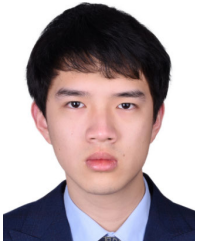

**Yuchao Shao** received the B.E. degree from Shanghai University of Engineering Science, Shanghai, China, in 2024, where he is currently pursuing the M.S. degree. His research interests include computer vision, intelligent transportation systems, and industry applications.

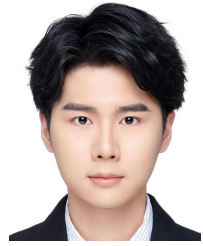

**Yaqi Liu** received the B.E. degree from Shanghai University of Engineering Science, Shanghai, China, in 2024, where he is currently pursuing the M.S. degree. His research interests include computer vision and robotics.

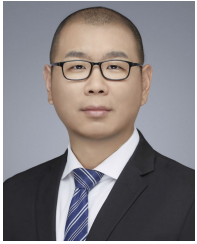

**Yang Zheng** (Member, IEEE) received the Ph.D. degree from the University of Science and Technology Beijing, China, in 2018.

He is currently an Associate Researcher at the Institute of Automation, Chinese Academy of Sciences. His research interests include computer vision and speech recognition.

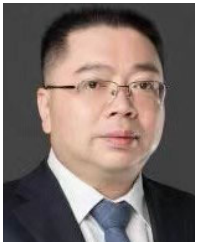

**Xiang Liu** (Member, IEEE) received the B.Sc. degree from Nanjing Normal University, the M.Eng. degree from Jiangsu University, and the Ph.D. degree from Fudan University. He is currently a Professor and the Director of the Department of Computer Science, School of Electronic and Electrical Engineering, Shanghai University of Engineering Science, Shanghai, China. His current research interests include computer vision and pattern recognition.

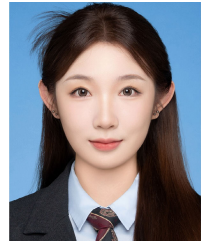

**Yitong Cai** received the B.E. degree from Liaoning Technical University, Huludao, China, in 2024. She is currently pursuing the M.S. degree with Shanghai University of Engineering Science. Her research interests include computer vision and robotics.